\documentclass{article}
\usepackage{spconf,amsmath,graphicx}
\usepackage[nolist]{acronym}
\usepackage{enumitem}
\usepackage{hyperref}
\begin{acronym}
\acro{RNN}{recurrent neural network}
\acro{RNNs}{recurrent neural networks}
\acro{GRU}{gated recurrent unit}
\acro{LSTM}{long short-term memory}
\acro{MFCC}{Mel-frequency cepstral coefficient}
\acro{MFCCs}{Mel-frequency cepstral coefficients}
\acro{PEM}{per-epoch noise mixing}
\acro{SNR}{signal-to-noise ratio}
\acro{ASR}{automatic speech recognition}
\acro{CNN}{convolutional neural network}
\acro{DNN}{deep neural network}
\acro{DNNs}{deep neural networks}
\acro{ACCAN}{accordion annealing}
\acro{WSJ}{Wall Street Journal}
\acro{CTC}{Connectionist Temporal Classification}
\acro{WFST}{Weighted Finite State Transducer}
\acro{WER}{word error rate}
\acro{LVCSR}{large-vocabulary continuous speech recognition}
\acro{ROI}{range of interest}
\end{acronym}

\usepackage{color}

\ninept

\title{A Curriculum Learning Method for Improved Noise Robustness in Automatic Speech Recognition}
\makeatletter
\def\name#1{\gdef\@name{#1\\}}
\makeatother \name{{\em Stefan Braun, Daniel Neil, and Shih-Chii Liu}}

\address{Institute of Neuroinformatics, University of Zurich and ETH Zurich\\
Zurich, Switzerland\\
  {\small \tt 	brauns@student.ethz.ch,neil@ini.uzh.ch,shih@ini.ethz.ch}
}

\begin{document}
%
\maketitle
  \begin{abstract}
The performance of \acl{ASR} systems under noisy environments still leaves room for improvement. Speech enhancement or feature enhancement techniques for increasing noise robustness of these systems usually add components to the recognition system that need careful optimization. In this work, we propose the use of a relatively simple curriculum training strategy called \ac{ACCAN}. It uses a multi-stage training schedule where samples at \ac{SNR} values as low as 0dB are first added and samples at increasing higher \ac{SNR} values are gradually added up to an \ac{SNR} value of 50dB. We also use a method called \ac{PEM} that generates noisy training samples online during training and thus enables dynamically changing the \ac{SNR} of our training data. Both the \ac{ACCAN} and the \ac{PEM} methods are evaluated on a end-to-end speech recognition pipeline on the Wall Street Journal corpus. \ac{ACCAN} decreases the average \ac{WER} on the 20dB to -10dB \ac{SNR} range by up to 31.4\% when compared to a conventional multi-condition training method. 

\end{abstract}

\section{Introduction}
\label{sec:intro}
The performance of \ac{ASR} systems has increased significantly with the use of \acp{DNN} \cite{hinton2012deep}. However, their performance in noisy environments still leaves room for improvement. Over the past decades, a multitude of methods to improve noise robustness of \ac{ASR} systems has been proposed \cite{li2014overview}, with many methods being applicable to \acp{DNN}. These methods enhance noise robustness at various levels and are applied prior to feature extraction, at the feature level, and during training.

Some example enhancement methods applied prior to feature extraction include denoising methods \cite{boll1979suppression} and source separation methods \cite{fujita2015unified} \cite{hori2015merl}. Methods applied at the feature level include methods that produce auditory-level features \cite{hermansky1991compensation} and feature space adaptation methods \cite{fmllr}. Other approaches use DNNs, e.g. feature denoising with deep autoencoders \cite{maas2012recurrent} \cite{feng2014speech} or feature extraction from the raw waveform via \acp{CNN} \cite{palaz2015analysis} \cite{adieu}. Many of these strategies add components to the speech recognition system that need careful optimization. 

The training method itself can have a major influence on the performance of a neural network under noisy conditions. Training on noisy data is an established method of increasing the noise robustness of a network. Noisy training sets with a range of \ac{SNR} values e.g. 10\,dB - 20\,dB \cite{seltzer2013investigation} or 0\,dB - 30\,dB \cite{palaz2015analysis} are used during training. Other training methods such as dropout \cite{dropout} - originally intended to improve regularisation - have been shown to also improve noise robustness \cite{seltzer2013investigation}. The same is true for model adaptation/noise aware training techniques \cite{seltzer2013investigation}. 

This paper presents general training methods for improving noise robustness in \ac{RNN}-based recognizers. \acp{RNN} are used here because they have demonstrated state-of-the-art performance in tasks such as the common sequence labelling task in speech recognition \cite{miao2015eesen} \cite{deepspeech2}. 

In particular, 
we introduce a new training strategy called {\bf \acf{ACCAN}} which exploits the benefits of curriculum-based training methods. By first training the network on low \ac{SNR} levels down to $0$\,dB  and gradually increasing the \ac{SNR} range to encompass higher \ac{SNR} levels, the trained network shows better noise robustness when tested under a wide range of \ac{SNR} levels.

This work also investigates the usefulness of adding noise both at the acoustic waveform level and at the feature representation level during training. In particular, we exploit a method called \textbf{\acf{PEM}}, which is a waveform-level data augmentation method. It enables us to generate a new training set for every epoch of training, i.e. each training sample is mixed with a newly sampled noise segment at a randomly chosen \ac{SNR} in every epoch. This form of data augmentation prevents networks from relying on constant noise segments for classification and helps in creating the necessary training samples over a wide \ac{SNR} range. These steps lead to improved generalization and noise robustness of the trained network. Our results are evaluated on the \textit{Wall Street Journal} corpus in a \ac{LVCSR} task. The testing is carried out on a large \ac{SNR} range going from clean conditions ($>$ 50\,dB) down to -20\,dB. 

The paper is organized as follows: Section \ref{sec:training} presents our training methods for improved noise robustness. The evaluation setup is detailed in Section \ref{sec:setup}, with results given in Section \ref{sec:results} followed by discussions in Section \ref{sec:discussion} and concluding remarks in Section \ref{sec:conclusions}.

\section{Training methods for improved noise robustness}
\label{sec:training}
\subsection{Baseline}
\label{sec:baseline}
Our baseline method takes advantage of multi-condition training \cite{yin2015noisy} in order to increase the noise robustness  of the network. Pink noise is added to a clean dataset to create samples with the desired \ac{SNR}. 
Each training sample is randomly chosen to be of an \ac{SNR} level in the range $0$ to $50$ dB with $5$ dB steps. This wide range is larger than the SNR ranges used in previous work (e.g. $0$ to $30$ dB as in \cite{palaz2015analysis}). Our exhaustive simulations show that using such a large range resulted in the best performance on the test datasets. The noise mixing is done once at the waveform-level before filterbank audio features are computed. This one set of training data is presented to the network over all training epochs. The resulting networks will be referred to as ``noisy-baseline''. For completeness, we also include a ``clean-baseline'', i.e. a network that is only trained on clean speech. 

\subsection{Gaussian noise injection}
\label{sec:gauss}
Gaussian noise injection is a well-known method for improving generalisation in neural networks \cite{koistinen1991kernel}. 
It is used here to improve the noise robustness of the network.

During training, artificial Gaussian noise is added to the filterbank features created from the different \ac{SNR} samples. The amount of additive noise is drawn from a zero-centered Gaussian distribution. Using a Gaussian with a standard deviation of $\sigma=0.6$ yielded the best results. This method is referred to as the ``Gauss-method'' in the rest of the paper.
    
\subsection{Per-epoch noise mixing (PEM)}
\label{sec:pem}
\ac{PEM} is a method for adding noise to the waveform level during training. In every training epoch, each training sample is mixed with a randomly sampled noise segment at a randomly sampled \ac{SNR}. The training procedure consists of the following steps:    
    \begin{enumerate}
    \item Mix every training sample with a randomly selected noise segment from a large pink noise pool to create a resulting sample at a randomly chosen \ac{SNR} level between $0$ to $50$ dB.
    \item Extract audio features (e.g. filterbank features) for the noise-corrupted audio to obtain the training data for the current epoch.
    \item \textit{Optional}: add Gaussian noise to the audio features.  
    \item Train on newly generated training data for one epoch.
    \item Discard this training data after the epoch to free up storage.
    \item Repeat from step 1 until training terminates.
    \end{enumerate}
    
This method has several key advantages over conventional pre-training preprocessing methods. Firstly, it enables unlimited data augmentation on large speech datasets. With conventional methods, augmenting training data at the waveform level with real-world noise at various \ac{SNR} values is prohibitively expensive in terms of processing time and training data size.  \ac{PEM} allows use to overcome these restrictions by training on the GPU and pre-processing the next-epoch training data in parallel on the CPU. After an epoch was trained on, the training data gets discarded to free storage for the next epoch.


Secondly, \ac{PEM} shows the network more unique training data: every training sample is presented at a selection of \acp{SNR} and with as many noise samples as can be extracted from the noise file and as needed by the number of epochs to reach a steady-state accuracy level. Thirdly, other noise types, different \ac{SNR} training ranges, and even different audio features could be quickly tested as the training data can be easily augmented online. Finally, \ac{PEM} enables us to dynamically change the \ac{SNR} level during training, which renders advanced training paradigms such as curriculum learning (section \ref{sec:curriculum}) feasible.


In contrast to the Gauss-method,
\ac{PEM} permits more control over the training data. Real-world noise is added to the acoustic waveform at controlled \acp{SNR}, ensuring that the training data corresponds to realistic noise corruption with results that can be evaluated. 
Of course, \ac{PEM} can be combined with Gaussian noise addition (optional step three in Section \ref{sec:pem}). We refer to \ac{PEM} without Gaussian noise injection as ``Vanilla-\ac{PEM}'' and to \ac{PEM} with Gaussian noise injection as ``Gauss-\ac{PEM}''.

\subsection{Curriculum learning}
\label{sec:curriculum}
Neural networks have been shown to optimize on the \ac{SNR} they are trained on \cite{yin2015noisy}. A network trained on clean conditions thus fares worse than a network trained on noisy conditions. Also, networks trained on a vast \ac{SNR} range generally do worse on a single \ac{SNR} than networks optimized for this specific \ac{SNR}. In order to achieve high accuracy under both high and low \ac{SNR} with a single network, we explored novel training paradigms based on curriculum learning. While curriculum learning has been used in image classification (scheduled denoising autoencoders, \cite{sched}) as well as speech recognition (SortaGrad \cite{deepspeech2}, a method for faster accuracy convergence), this is the first work targeted at \ac{LVCSR} under noisy conditions. 

Our novel \ac{ACCAN} training method applies a multi-stage training schedule: in the first stage, the neural network is trained on the lowest \ac{SNR} samples. In the following stages, the \ac{SNR} training range is expanded in 5\,dB steps towards higher \ac{SNR} values. A typical schedule is shown in \autoref{annealing}. In every stage, training repeats until the \ac{WER} on the development set no longer improves. At the end of each stage, the weights of the best network are stored and used as the starting point for the next stage. Both training and validation sets share the same \ac{SNR} range. The \ac{ACCAN} approach seems counter-intuitive as noisy training data should be harder to train on than clean data. However, the noise allows the network to explore the parameter space more extensively at the beginning \cite{curriculum}. We also evaluated a method called ``\ac{ACCAN}-reversed'', which expands from high \ac{SNR} to low \ac{SNR}, but the results were very close to the standard ``Gauss-PEM'' approach.

\begin{table}
\centering
\caption{The \ac{ACCAN} training strategy: \ac{SNR} ranges [dB] of the training stages}
\begin{tabular}{|c|c|c|c|c|}
\hline 
Method & Stage 1 & Stage 2 & ... & Stage 10 \\ 
\hline 
\ac{ACCAN} & [0] & [0, 5] & ... & [0, 5, ..., 45, 50]  \\ 
\hline 
\ac{ACCAN} reversed & [50] & [50, 45] & ... & [50, 45, ..., 5, 0]  \\ 
\hline
\end{tabular} 
\label{annealing}
\end{table}

\section{Setup}
\label{sec:setup}
\paragraph*{Audio database}
All experiments were carried out on the \ac{WSJ} corpus (LDC93S6B and LDC94S13B) in the following configuration: 
\begin{itemize}
\item training set: \textit{train-si84} (7138 samples),
\item development set: \textit{test-dev93} (503 samples),
\item test set: \textit{test-eval92} (333 samples).
\end{itemize}
For noise corruption, we used two different noise types: \textit{pink noise} generated by the Audacity \cite{audacity} software and \textit{babble noise} from the NOISEX database \cite{noisex92}.

\paragraph*{Data preparation and language model}
The labels and transcriptions were extracted with EESEN \cite{miao2015eesen} routines. All experiments were character-based and used 58 labels (letters, digits, punctuation marks etc.) During test time, the network output was decoded with the \ac{WFST} approach from the EESEN framework, which allows us to apply a trigram language model. The language model used an expanded vocabulary in order to avoid out-of-vocabulary words occurring in the standard \ac{WSJ} language model.

\paragraph*{Audio features}
We used 123-dimensional filterbank features that consisted of 40 filterbanks, 1 energy term and their respective first and second order derivatives. The features were generated by preprocessing routines from EESEN \cite{miao2015eesen}. Each feature dimension is zero-mean and unit-variance normalized.

\paragraph*{Neural network configuration}
Our recognition pipeline is a end-to-end solution that involves a \ac{RNN} as the acoustic model. In order to automatically learn the alignments between speech frames and label sequences, the \ac{CTC} \cite{graves2006connectionist} objective was adopted. The Lasagne library \cite{lasagne} enabled us to build and train our 5-layer neural network. The first 4 layers consisted of bidirectional \ac{LSTM} \cite{hochreiter1997long} units with 250 units in each direction. The fifth and final layer was a non-flattening dense layer with 59 outputs, corresponding to the character labels + the blank label required by CTC. The network contained 8.5M tunable parameters. All layers were initialized with the Glorot uniform strategy \cite{glorot}. Every experiment started with the exact same weight initialization. During training, the Adam \cite{adam} stochastic optimization method was used. To prevent overfitting and to increase noise robustness, dropout \cite{dropout} was used (dropout probability=0.3). Every epoch of training, the \ac{WER} on the development set was monitored with a simple best-path decoding approach.

With all training strategies except \ac{ACCAN}, the network was trained for a generous 150 epochs.  The networks weights from the epoch with the lowest \ac{WER} were kept for evaluation. Generally, the improvements in \ac{WER} saturated well before 150 epochs were reached. The \ac{ACCAN} method used a patience of 5 as to switch between \ac{SNR} stages, i.e. if the \ac{WER} did not improve for 5 epochs on the current \ac{SNR} stage, the training continued on the next \ac{SNR} stage. By respecting the stage-switching policy, \ac{ACCAN} reached the final \ac{SNR} stage with the full \ac{SNR} range at epoch 190. Saturation kicked in at epoch 240. While \ac{ACCAN} trained for more epochs than the others, it only trained for 50 epochs on the full \ac{SNR} range.

\section{Results}
\label{sec:results}
The reported results are given for the 'test-eval92' evaluation set for the \textit{Wall Street Journal} corpus. The evaluation set was tested in clean condition and with added pink noise or babble noise at 15 \ac{SNR} levels from 50dB to -20 dB in dB steps. We report in \autoref{compact}  the average \ac{WER} over following \ac{SNR} ranges:
\begin{itemize}
\item Full \ac{SNR} range: [clean signal, 50dB to -10dB]
\item High \ac{SNR} range: [50dB to 0dB]
\item Low  \ac{SNR} range: [0dB to -10dB]
\item \ac{ROI}: [20dB to -10dB]
\end{itemize}
We choose to include the \ac{ROI}, as our hearing tests showed that this range seems to well reflect common scenarios in public environments, where a clean speech signal is most often not found. Detailed results for each \ac{SNR} individually are given in \autoref{detailed}. Results for -15dB and -20dB are reported too, but should be considered as extreme cases. \ac{WER} improvements are given as relative improvements in the text.

\begin{table}[htbp]
  \centering
  \caption{Average absolute \ac{WER} [\%] for given \ac{SNR} ranges after decoding. Printed bold: lowest \ac{WER}.}
    \begin{tabular}{|l|c|c|c|c|}
    \hline
    \multicolumn{5}{|c|}{\textbf{Testing against pink noise}} \\
    \hline
    Method & Full  & High  & Low   & ROI \\
    \hline
    Clean-baseline & 54.7  & 29.0  & 109.6 & 67.9 \\
    Noisy-baseline & 46.0  & 23.3  & 88.6  & 51.7 \\
    Gauss & 37.4  & 19.8  & 71.1  & 42.1 \\
    Vanilla-PEM   & 35.6  & 17.8  & 70.6  & 40.8 \\
    Gauss-PEM & \textbf{34.1}  & \textbf{16.6} & 64.7  & 37.2 \\
    ACCAN & 34.4  & 18.1  & \textbf{59.5} & \textbf{36.0} \\
    ACCAN reversed & 35.2 & 17.8 & 66.3 & 38.8\\
    \hline
    \hline
    \multicolumn{5}{|c|}{\textbf{Testing against babble noise}} \\
    \hline
    Method & Full  & High  & Low   & ROI \\
    \hline
    Clean-baseline & 53.0  & 32.0  & 113.7 & 72.1 \\
    Noisy-baseline & 53.3  & 29.9  & 114.0 & 68.4 \\
    Gauss & 45.4  & 25.4  & 96.3  & 56.9 \\
    Vanilla-PEM   & 41.0  & 22.8  & 88.3  & 52.3 \\
    Gauss-PEM & \textbf{39.5}  & 21.6  & 83.7  & 49.0 \\
    ACCAN & 39.6  & \textbf{21.5} & \textbf{80.2} & \textbf{47.0} \\
    ACCAN reversed & \textbf{39.5}  & \textbf{21.5}  & 82.9 & 48.2 \\
    \hline
    \end{tabular}%
  \label{compact}%
\end{table}%

\begin{table*}[htbp]
\footnotesize
  \centering
  \caption{Absolute WER [\%] on single SNRs after decoding. Printed bold: lowest WER.}
    \begin{tabular}{|l|c|c|c|c|c|c|c|c|c|c|c|c|c|c|c|c|}
    \hline
    \multicolumn{17}{|c|}{\textbf{Testing against pink noise}} \\
    \hline
    \multicolumn{1}{|c|}{Method / SNR[dB]} & clean & 50.0  & 45.0  & 40.0  & 35.0  & 30.0  & 25.0  & 20.0  & 15.0  & 10.0  & 5.0   & 0.0   & -5.0  & -10.0 & -15.0 & -20.0 \\
    \hline
    Clean-baseline & 13.8  & 14.4  & 14.0  & 13.8  & 13.7  & 13.7  & 16.1  & 18.9  & 25.9  & 40.1  & 61.8  & 86.4  & 109.0 & 133.4 & 147.2 & 152.8 \\
    Noisy-baseline & 17.3  & 17.4  & 17.3  & 16.9  & 16.5  & 16.4  & 16.2  & 16.8  & 19.0  & 23.4  & 36.5  & 59.8  & 90.0  & 116.2 & 126.7 & 129.5 \\
    Gauss & 15.7  & 15.8  & 15.7  & 15.6  & 14.8  & 14.4  & 14.5  & 15.3  & 16.9  & 20.2  & 28.9  & 45.5  & 72.8  & 94.9  & 99.0  & 98.9 \\
    Vanilla-PEM   & \textbf{13.3} & \textbf{13.2} & \textbf{13.2} & \textbf{12.7} & \textbf{12.6} & \textbf{12.6} & 12.9  & 13.6  & 15.1  & 18.9  & 26.2  & 45.0  & 73.5  & 93.4  & 96.9  & 97.1 \\
    Gauss-PEM & 13.6  & 13.5  & 13.6  & 13.4  & 13.2  & \textbf{12.6} & \textbf{12.4} & \textbf{12.8} & \textbf{14.2} & \textbf{17.0} & \textbf{22.3} & 37.6  & 66.3  & 90.2  & 95.9  & 96.8 \\
    ACCAN & 15.9  & 15.8  & 15.4  & 15.3  & 15.0  & 15.0  & 15.2  & 15.9  & 16.1  & 18.5  & 22.9  & \textbf{33.7} & \textbf{58.8} & \textbf{85.9} & \textbf{95.6} & \textbf{96.2} \\
    ACCAN reversed & 14.6  & 14.4  & 14.3  & 14.1  & 13.3  & 13.4  & 13.9  & 14.4  & 15.4  & 18.5  & 24.4  & 40.2  & 67.8  & 90.9  & 97.0  & 97.2 \\
    \hline
    \hline
    \multicolumn{17}{|c|}{\textbf{Testing against babble noise}} \\
    \hline
    \multicolumn{1}{|c|}{Method / SNR[dB]} & clean & 50.0  & 45.0  & 40.0  & 35.0  & 30.0  & 25.0  & 20.0  & 15.0  & 10.0  & 5.0   & 0.0   & -5.0  & -10.0 & -15.0 & -20.0 \\
    \hline
    Clean-baseline & 13.8  & 14.2  & 14.2  & 13.9  & 14.2  & 14.5  & 15.7  & 18.8  & 26.6  & 43.9  & 74.2  & 102.2 & 116.6 & 122.4 & 122.3 & 121.4 \\
    Noisy-baseline & 17.3  & 17.1  & 16.9  & 16.7  & 16.1  & 15.7  & 15.8  & 17.8  & 23.1  & 35.5  & 60.6  & 94.1  & 119.4 & 128.4 & 129.3 & 129.2 \\
    Gauss & 15.7  & 15.6  & 15.7  & 15.7  & 15.3  & 15.0  & 15.4  & 16.5  & 19.5  & 27.5  & 45.9  & 77.4  & 102.6 & 109.0 & 109.8 & 110.6 \\
    Vanilla-PEM   & \textbf{13.3} & \textbf{13.2} & \textbf{12.9} & \textbf{12.7} & \textbf{12.3} & \textbf{12.7} & \textbf{12.8} & \textbf{14.0} & 17.4  & 25.6  & 44.2  & 72.7  & 93.2  & 99.1  & 99.5  & 99.8 \\
    Gauss-PEM & 13.6  & 13.8  & 13.7  & 13.4  & 13.4  & 13.3  & 13.7  & 14.6  & 16.9  & 22.9  & 37.4  & 64.1  & 89.8  & 97.3  & \textbf{97.3} & \textbf{97.4} \\
    ACCAN & 15.9  & 15.7  & 15.3  & 14.9  & 15.1  & 15.1  & 15.0  & 15.5  & 17.5  & \textbf{21.8} & \textbf{33.4} & \textbf{57.2} & \textbf{86.1} & 97.2  & 98.8  & 99.1 \\
    ACCAN reversed & 14.6  & 14.4  & 14.2  & 14.0  & 14.1  & 14.0  & 14.0  & 14.6  & \textbf{16.5} & 21.9  & 35.5  & 63.5  & 88.7  & \textbf{96.6} & 97.6  & 97.6 \\
    \hline
    \end{tabular}%
  \label{detailed}%
\end{table*}%

\subsection{Noise addition methods}
This section summarizes results from the \textit{baseline}, \textit{Gauss}, \textit{Vanilla-PEM} and \textit{Gauss-PEM} methods, all trained on the \ac{SNR} range from 0dB to 50dB. Our network trained on clean speech only (clean-baseline) achieves 13.8\% \ac{WER} with a trigram language model and our 8.5M parameter network, while in literature \cite{graves2014towards}, a 13.5\% \ac{WER} was achieved using a trigram language model and ~3x larger (26.5M) parameter network. This confirms that our end-to-end speech recognition pipeline is fully functional.

\paragraph*{Baseline:} The noisy-baseline network starts with a 25\% higher \ac{WER} on the clean test set than our clean-baseline network. For \acp{SNR} lower than 25dB, the noisy-baseline is significantly more noise robust. The \ac{WER} seems to drastically increase at 25dB for the clean-baseline, while the noisy-baseline sees the increase at a lower 10dB \ac{SNR}. However, all other methods outperform the noisy-baseline by a significant margin at high and low SNRs.

\paragraph*{Vanilla-PEM vs Gauss:} Compared to the noisy-baseline, Vanilla-PEM achieves a ~23\% decrease in \ac{WER} on high SNR, while Gauss only reduces \ac{WER} by 15\% (both pink noise and babble noise). This results in vanilla-PEM being able to outperform the clean-baseline on clean speech, while Gauss is not able to do the same. On low \ac{SNR}, both methods reduce \ac{WER} by around 20\% on the pink noise test set. On babble noise, \ac{PEM} results in a higher 22.5\% WER decrease compared to the 15.5\% decrease provided by Gauss.

\paragraph*{Gauss-PEM:} The Gauss-PEM method achieves the overall lowest \ac{WER} on the high and low SNR range. It beats the noisy-baseline method by between 26.5\% and 28.7\% on high SNR, on low SNR and on the \ac{ROI} for both pink noise and babble noise. The results on the high SNR range are notable: Gauss-PEM is able to outperform the clean-baseline network at every single \ac{SNR} step in the high \ac{SNR} range, even on clean speech. The network is much more noise robust while at the same time it even improves clean speech scores.
Around 35dB to 25dB, Gauss-PEM (other methods, too) reaches its minimum \ac{WER}. This is expected, as the mean \ac{SNR} of the training \ac{SNR} range is 25dB and the network seems to optimize for \ac{SNR} levels close to this value \cite{yin2015noisy}.

\subsection{Curriculum learning}
To further increase the noise robustness, we developed a curriculum learning strategy for the Gauss-PEM method, resulting in our novel \ac{ACCAN} method. We compare our results to Gauss-PEM, as this was the most noise robust non-curriculum method. Our test results show increased noise robustness for \ac{ACCAN} on pink noise and babble noise: the \ac{WER} decreases between 3.3\% (ROI, pink noise) and 4.1\% (ROI babble noise). For pink noise, the biggest decrease is seen at 0dB  (10.5\% \ac{WER} decrease) and -5dB (11.3\% decrease). For babble noise, the biggest \ac{WER} decreases are found at 10dB (4.9\%), 5dB (10.9\%) and 0dB (10.7\%). 

The average \ac{WER} of \ac{ACCAN} on the high \ac{SNR} range is worse on pink noise (relative 8.8\% increase in \ac{WER}), but better on babble noise (relative 0.4\% decrease in \ac{WER}). Ultimately, the absolute \ac{WER} in clean speech of \ac{ACCAN} (15.9\%) is better than the noisy-baseline (17.3\%) but worse than Gauss-PEM (13.6\%).

\section{Discussion}
\label{sec:discussion}

All proposed training methods lead to networks with increased noise robustness in the low \ac{SNR} range in comparison with the standard noisy-baseline method. The noise robustness is increased on a network level and it does not rely on complex preprocessing frameworks. 

We see increased noise robustness against both tested noise types. This is remarkable, as the networks only saw the pink noise type during training. The results show that waveform-level noise mixing (\ac{PEM}) is especially strong in transferring noise robustness to noise types not seen in training. The feature level noise addition (Gauss) is less effective on unseen noise types. Also, \ac{PEM} enabled us to train noise robust networks that - at the same time - achieve lower \ac{WER} on clean speech than a network trained only on clean speech. The uncompromising data augmentation by \ac{PEM} should be a decisive factor to achieve these results. While the noisy-baseline trained on 1.7GB of unique data (waveform level), the \ac{PEM}-enabled networks trained on up to 408GB (240 epochs for ACCAN * 1.7GB) of unique training data (waveform level). By permanently sampling different noise segments, we force the network to not rely only on constant noise features for classification, but to develop a better internal representation of the speech data. This representation could be refined further by using other noise types besides pink noise for training, such as babble noise, street noise, restaurant noise. Also, \ac{SNR} steps smaller than 5dB could be used to allow more than 11 different \ac{SNR} levels during training.

\ac{PEM} enabled us to dynamically change the \ac{SNR} during training. This facilitated the implementation of our novel \ac{ACCAN} training strategy, that achieved the best noise robustness performance. The multi-stage training starts at low \acp{SNR}, where annealed networks are able to explore the parameters space with moderate influence of the speech signal. During gradual exposure to higher \acp{SNR} in the training process, accordion annealed networks refine their internal model of speech step by step, while they seemingly acquire higher noise robustness at the lower \ac{SNR} levels. The inverse way of going through the \ac{SNR} range does not yield increased noise robustness. The immediate presence of clean speech signals seems to force the network to converge faster to a complex acoustic model instead of exploring the parameter space.


\section{Conclusion}
\label{sec:conclusions}

This work proposes new training methods for improving the noise robustness of \acp{RNN} for a \ac{LVCSR} task. The networks are trained for a wide SNR range with the use of the Vanilla-PEM training method which adds noise at the waveform level and the Gauss method which injects Gaussian noise at the feature level.
By combining the Gauss and Vanilla-PEM methods into the Gauss-PEM method, we achieve on average a 28\% \ac{WER} reduction on the 20dB to -10 dB SNR range when compared to a conventional multi-condition training method. At the same time, we achieve lower \ac{WER} on clean speech than a network that is trained solely on clean speech.
The \ac{ACCAN} training strategy enhances the Gauss-PEM method with a curriculum learning strategy. The \ac{ACCAN} training strategy results in performance up to 11.3\% lower \ac{WER} at low \ac{SNR}s compared to Gauss-PEM method.

\section{Acknowledgements}
\label{sec:acknow}

The authors acknowledge Enea Ceolini and Joachim Ott for discussions on RNNs. They also acknowledge Ying Zhang and Yoshua Bengio (both from University of Montreal) for help in the setup of the language model. This work was partially supported by Samsung Advanced Institute of Technology and EU H2020 COCOHA \#644732. 

\vfill\pagebreak



\bibliographystyle{IEEEbib}
\bibliography{strings,refs}

\begin{thebibliography}{10}

\bibitem{hinton2012deep}
Geoffrey Hinton, Li~Deng, Dong Yu, George~E Dahl, Abdel-rahman Mohamed, Navdeep
  Jaitly, Andrew Senior, Vincent Vanhoucke, Patrick Nguyen, Tara~N Sainath,
  et~al.,
\newblock ``Deep neural networks for acoustic modeling in speech recognition:
  The shared views of four research groups,''
\newblock {\em IEEE Signal Processing Magazine}, vol. 29, no. 6, pp. 82--97,
  2012.

\bibitem{li2014overview}
Jinyu Li, Li~Deng, Yifan Gong, and Reinhold Haeb-Umbach,
\newblock ``An overview of noise-robust automatic speech recognition,''
\newblock {\em IEEE/ACM Transactions on Audio, Speech, and Language
  Processing}, vol. 22, no. 4, pp. 745--777, 2014.

\bibitem{boll1979suppression}
Steven~F Boll,
\newblock ``Suppression of acoustic noise in speech using spectral
  subtraction,''
\newblock {\em IEEE Transactions on Acoustics, Speech and Signal Processing},
  vol. 27, no. 2, pp. 113--120, 1979.

\bibitem{fujita2015unified}
Yusuke Fujita, Ryoichi Takashima, Takeshi Homma, Rintaro Ikeshita, Yohei
  Kawaguchi, Takashi Sumiyoshi, Takashi Endo, and Masahito Togami,
\newblock ``Unified asr system using lgm-based source separation, noise-robust
  feature extraction, and word hypothesis selection,''
\newblock in {\em 2015 IEEE Workshop on Automatic Speech Recognition and
  Understanding (ASRU)}, 2015, pp. 416--422.

\bibitem{hori2015merl}
Takaaki Hori, Zhuo Chen, Hakan Erdogan, John~R Hershey, JL~Roux, Vikramjit
  Mitra, and Shinji Watanabe,
\newblock ``The merl/sri system for the 3rd chime challenge using beamforming,
  robust feature extraction, and advanced speech recognition,''
\newblock in {\em Proc. IEEE ASRU}, 2015.

\bibitem{hermansky1991compensation}
Hynek Hermansky, Nelson Morgan, Aruna Bayya, and Phil Kohn,
\newblock ``Compensation for the effect of the communication channel in
  auditory-like analysis of speech (rasta-plp),''
\newblock in {\em Second European Conference on Speech Communication and
  Technology}, 1991.

\bibitem{fmllr}
Takaaki Hori, Zhuo Chen, Hakan Erdogan, John~R Hershey, JL~Roux, Vikramjit
  Mitra, and Shinji Watanabe,
\newblock ``The merl/sri system for the 3rd chime challenge using beamforming,
  robust feature extraction, and advanced speech recognition,''
\newblock in {\em Proc. IEEE ASRU}, 2015.

\bibitem{maas2012recurrent}
Andrew~L Maas, Quoc~V Le, Tyler~M O'Neil, Oriol Vinyals, Patrick Nguyen, and
  Andrew~Y Ng,
\newblock ``Recurrent neural networks for noise reduction in robust asr.,''
\newblock in {\em INTERSPEECH}, 2012, pp. 22--25.

\bibitem{feng2014speech}
Xue Feng, Yaodong Zhang, and James Glass,
\newblock ``Speech feature denoising and dereverberation via deep autoencoders
  for noisy reverberant speech recognition,''
\newblock in {\em 2014 IEEE International Conference on Acoustics, Speech, and
  Signal Processing (ICASSP)}. IEEE, 2014, pp. 1759--1763.

\bibitem{palaz2015analysis}
Dimitri Palaz, Ronan Collobert, et~al.,
\newblock ``Analysis of cnn-based speech recognition system using raw speech as
  input,''
\newblock in {\em Proceedings of Interspeech}, 2015.

\bibitem{adieu}
George Trigeorgis, Fabien Ringeval, Raymond Brueckner, Erik Marchi, Mihalis~A
  Nicolaou, Stefanos Zafeiriou, et~al.,
\newblock ``Adieu features? end-to-end speech emotion recognition using a deep
  convolutional recurrent network,''
\newblock in {\em 2016 IEEE International Conference on Acoustics, Speech and
  Signal Processing (ICASSP)}. IEEE, 2016, pp. 5200--5204.

\bibitem{seltzer2013investigation}
Michael~L Seltzer, Dong Yu, and Yongqiang Wang,
\newblock ``An investigation of deep neural networks for noise robust speech
  recognition,''
\newblock in {\em 2013 IEEE International Conference on Acoustics, Speech, and
  Signal Processing (ICASSP)}. IEEE, 2013, pp. 7398--7402.

\bibitem{dropout}
Nitish Srivastava, Geoffrey Hinton, Alex Krizhevsky, Ilya Sutskever, and Ruslan
  Salakhutdinov,
\newblock ``Dropout: A simple way to prevent neural networks from
  overfitting,''
\newblock {\em The Journal of Machine Learning Research}, vol. 15, no. 1, pp.
  1929--1958, 2014.

\bibitem{miao2015eesen}
Yajie Miao, Mohammad Gowayyed, and Florian Metze,
\newblock ``Eesen: End-to-end speech recognition using deep rnn models and
  wfst-based decoding,''
\newblock in {\em 2015 IEEE Workshop on Automatic Speech Recognition and
  Understanding (ASRU)}, 2015, pp. 167--174.

\bibitem{deepspeech2}
Dario Amodei et~al.,
\newblock ``Deep speech 2: End-to-end speech recognition in english and
  mandarin,''
\newblock {\em CoRR}, vol. abs/1512.02595, 2015.

\bibitem{yin2015noisy}
Shi Yin, Chao Liu, Zhiyong Zhang, Yiye Lin, Dong Wang, Javier Tejedor,
  Thomas~Fang Zheng, and Yinguo Li,
\newblock ``Noisy training for deep neural networks in speech recognition,''
\newblock {\em EURASIP Journal on Audio, Speech, and Music Processing}, vol.
  2015, no. 1, pp. 1--14, 2015.

\bibitem{koistinen1991kernel}
Petri Koistinen and Lasse Holmstr{\"o}m,
\newblock ``Kernel regression and backpropagation training with noise,''
\newblock in {\em 1991 IEEE International Joint Conference on Neural Networks},
  1991, pp. 367--372.

\bibitem{sched}
Krzysztof~J. Geras and Charles~A. Sutton,
\newblock ``Scheduled denoising autoencoders,''
\newblock {\em CoRR}, vol. abs/1406.3269, 2014.

\bibitem{curriculum}
Yoshua Bengio, J{\'e}r\^{o}me Louradour, Ronan Collobert, and Jason Weston,
\newblock ``Curriculum learning,''
\newblock in {\em Proceedings of the 26th Annual International Conference on
  Machine Learning}, New York, NY, USA, 2009, ICML '09, pp. 41--48, ACM.

\bibitem{audacity}
Audacity team,
\newblock ``Audacity,'' 2016.

\bibitem{noisex92}
Andrew Varga and Herman~JM Steeneken,
\newblock ``Assessment for automatic speech recognition: Ii. noisex-92: A
  database and an experiment to study the effect of additive noise on speech
  recognition systems,''
\newblock {\em Speech Communication}, vol. 12, no. 3, pp. 247--251, 1993.

\bibitem{graves2006connectionist}
Alex Graves, Santiago Fern{\'a}ndez, Faustino Gomez, and J{\"u}rgen
  Schmidhuber,
\newblock ``Connectionist temporal classification: labelling unsegmented
  sequence data with recurrent neural networks,''
\newblock in {\em Proceedings of the 23rd International Conference on Machine
  learning}. ACM, 2006, pp. 369--376.

\bibitem{lasagne}
Lasagne team,
\newblock ``Lasagne: First release.,'' Aug. 2015.

\bibitem{hochreiter1997long}
Sepp Hochreiter and J{\"u}rgen Schmidhuber,
\newblock ``Long short-term memory,''
\newblock {\em Neural Computation}, vol. 9, no. 8, pp. 1735--1780, 1997.

\bibitem{glorot}
Xavier Glorot and Yoshua Bengio,
\newblock ``Understanding the difficulty of training deep feedforward neural
  networks,''
\newblock in {\em International Conference on Artificial Intelligence and
  Statistics}, 2010, pp. 249--256.

\bibitem{adam}
Diederik~P. Kingma and Jimmy Ba,
\newblock ``Adam: {A} method for stochastic optimization,''
\newblock {\em CoRR}, vol. abs/1412.6980, 2014.

\bibitem{graves2014towards}
Alex Graves and Navdeep Jaitly,
\newblock ``Towards end-to-end speech recognition with recurrent neural
  networks.,''
\newblock in {\em ICML}, 2014, vol.~14, pp. 1764--1772.

\end{thebibliography}

\end{document}